\pdfoutput=1
\documentclass[11pt]{article}

\usepackage{acl}

\usepackage{times}
\usepackage{latexsym}
\usepackage{amsfonts}
\usepackage[T1]{fontenc}
\usepackage{booktabs}
\usepackage{xcolor}   
\usepackage{color}
\usepackage{booktabs}
\usepackage{amsmath}
\usepackage{amssymb}
\usepackage{colortbl}
\usepackage{tabularx}
\usepackage{multirow}
\usepackage{makecell}
\usepackage{subfig}
\usepackage{xspace}
\usepackage{booktabs}       
\usepackage{microtype}
\usepackage{inconsolata}
\usepackage{longtable}
\usepackage{graphicx}
\usepackage{colortbl}
\usepackage{hyphenat}
\usepackage[utf8]{inputenc}
\usepackage{CJK}
\usepackage{algorithm}
\usepackage{algpseudocode}

\usepackage{etoolbox}
\usepackage{pgf} 
\usepackage{xcolor, pgfmath}
\usepackage{etoolbox}

\definecolor{highcolor}{HTML}{44ff44} 
\definecolor{midcolor}{HTML}{ffffff}  
\definecolor{lowcolor}{HTML}{ff0000}  

\title{Improving Multilingual Language Models by Aligning Representations through Steering}

\author{
  Omar Mahmoud$^{\ast}$, Buddhika Laknath Semage$^{\dagger}$, Thommen George Karimpanal$^{\ddagger}$, and Santu Rana$^{\ast}$ \\
  $^{\ast}$Applied Artificial Intelligence Initiative, Deakin University, Australia \\
  $^{\ddagger}$School of Information Technology, Deakin University, Australia \\
  $^{\dagger}$Independent \\
  \texttt{o.mahmoud@deakin.edu.au}
}

\begin{document}

\maketitle

\begin{abstract}
This paper investigates how Large Language Models (LLMs) represent non-English tokens—a question that remains underexplored despite recent progress. We propose a lightweight intervention method using \textit{representation steering}, where a learned vector is added to the residual stream at a single model layer to enhance multilingual performance. Through extensive experiments across seven competitive baselines—including prompt optimization, supervised fine-tuning (\textsc{SFT}), in-context learning, cross-lingual transfer, and translation-based methods—we show that our approach consistently outperforms most alternatives. In particular, it achieves performance on par with production-grade translation systems while requiring far fewer resources. We further explore the complementarity between our method and \textsc{SFT}, demonstrating that steering offers a direct, efficient way to realign internal representations. These findings underscore the potential of activation-level interventions as a powerful tool for improving the multilingual capabilities of LLMs.
\end{abstract}

\section{Introduction}
In recent years, large language models (LLMs) have demonstrated remarkable capabilities across a wide range of tasks. However, the majority of these advancements have been concentrated in English, often neglecting other languages, particularly low-resource ones, due to the scarcity of available data. A common approach to addressing this gap is translating these languages into English before processing them. While this method can be effective, it is inherently limited by the quality and cost of translation~\cite{liu2024translation}. To unlock the full potential of LLMs, integrating multilingual natively within these models is essential, ensuring robust performance across diverse languages without relying solely on translation.

\noindent Recent studies have explored various strategies to enhance the multilingual proficiency of LLMs. These include cross-lingual fine-tuning \cite{qin-etal-2023-cross}, instruction alignment via code-switching \cite{huang2023not}, and chain-of-thought prompting in multiple languages \cite{shi2022language}. Other approaches focus on mapping representations between models, such as MindMerger, which integrates an external LLM's linguistic knowledge by learning a mapping between representation spaces \cite{huang2024mindmerger}. Despite these advances, investigations into the internal mechanisms of multilingual processing remain limited. The few studies in this area suggest that LLMs often default to translating non-English inputs into English representations within their intermediate layers \cite{wendler2024llamas, zhao2024largelanguagemodelshandle}.

\begin{figure*}[t]
\begin{center}
  \includegraphics[width=0.99\textwidth]{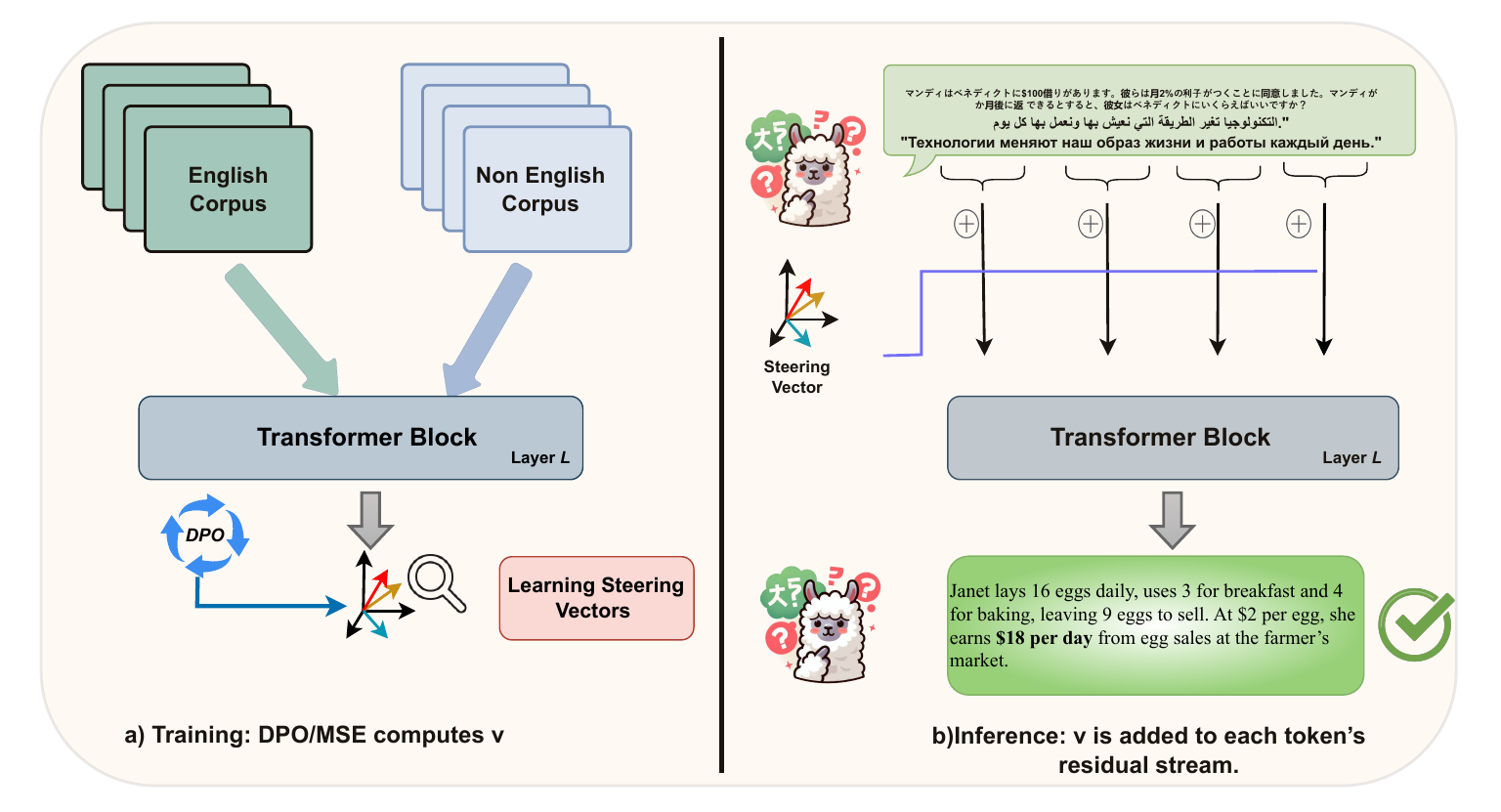}
      \vskip -0.3cm

  \caption{Overview of our method: (a) Learn a steering vector $v$ from two language corpora at a specific layer using DPO or MSE; (b) Apply the learned vector to the residual stream of each token in a prompt at that layer.
}
  \label{fig:main_diagram}
\end{center}

\end{figure*}

\noindent Closely related to our work, recent research \cite{wang2024bridging} has used representation steering to align hidden states between English and a target language. This was achieved by learning a steering vector through a least-squares optimization and applying it across all layers of the model. While effective, this method modifies the model's representations globally. This leaves a critical question unexplored: \textit{can multilingual alignment be achieved more efficiently by targeting only a single, specific layer, and what does this reveal about the model's internal structure?}
\noindent In this paper, we investigate this question from a mechanistic interpretability perspective. We propose a method that first learns a transformation manifold mapping English representations to a target language. This mapping is then applied as a steering vector to the activations of only a single layer during inference, without any fine-tuning. This lightweight approach is more efficient and less disruptive to the base model's capabilities, as illustrated in \autoref{fig:main_diagram}. By demonstrating that our single-layer intervention parallels the effects of full fine-tuning, we provide new insights into how LLM representations can be precisely and efficiently optimized for multilingual tasks.
Our key contributions are as follows:

\begin{itemize}

\item We propose and validate a method to enhance the multilingual capabilities of LLMs by steering the representations of a single layer, using a learned alignment with English.

\item We demonstrate that a single steering vector can be shared across structurally similar languages \footnote{Structurally similar languages share features—genetic, geographic, syntactic, phonological, featural, and inventory-based—as defined by the \href{https://github.com/antonisa/lang2vec}{lang2vec} framework.} , enabling zero-shot cross-linguistic transfer without language-specific fine-tuning.

\item Our method significantly surpasses the performance of the NLLB translation baseline and achieves results competitive with Google Translate across multiple datasets.
\end{itemize}

\section{Related Work}
\textbf{Multilingual Progress:} Recent research has significantly advanced multilingual LLMs, as highlighted in a survey by \cite{qin2024multilingual}. Efforts to enhance multilingual performance primarily focus on expanding language coverage through cross-lingual instruction fine tuning. For example, \cite{zhu2023extrapolating} and \cite{chen2023monolingual} propose multilingual instruction tuning methods to improve reasoning across diverse languages, while \cite{zhu2024question} integrates mathematical instructions to enhance logical processing. Another line of work explores prompt-based strategies to strengthen cross-lingual understanding. Studies by \cite{qin-etal-2023-cross,huang2023not} show that strategically designed prompts can significantly enhance model performance across languages. More recent methods introduce external modules to supplement the model's multilingual capabilities. \cite{yoon2024langbridge} propose LangBridge, which integrates a multilingual encoder with an LLM for improved reasoning, though it may underutilize the LLM’s native multilingual abilities, in contrast, MindMerger \cite{huang2024mindmerger} aligns representations across models handling the same prompt, preserving intrinsic multilingual features. Despite these advances, fewer studies focus on how LLMs internally manage multilingualism. Notably, \cite{wendler2024llamas} and \cite{zhao2024largelanguagemodelshandle} analyze the internal mechanisms enabling cross-lingual understanding, highlighting both strengths and limitations that inform further improvements.
\\ 
\textbf{Representation Engineering}
has emerged as a powerful tool for analyzing how concepts are processed within LLMs, addressing challenges such as truthfulness, fairness, and model editing \cite{zou2023representation}. This approach has been used to enhance model alignment and detect vulnerabilities, including jailbreaking risks in open-source models \cite{wang2024trojanactivationattackredteaming,li2024open}. Additionally, studies have leveraged it to investigate how LLMs internally represent complex concepts \cite{lu2024investigatingbiasrepresentationsllama}. Recent work by \cite{cao2024personalized} presents methods to extract refined steering vectors through preference optimization, allowing improved control of model behavior. These findings underscore the significant role of representation engineering in advancing LLM technology.\\
\noindent \textbf{Inference Time Intervention}: using steering vectors is an established technique in the field of model editing \cite{li2023inference, panickssery2023steering}. These methods modify model behavior by directly manipulating internal states; for example, vectors can be added to a model's residual stream to improve truthfulness \cite{wang2025adaptive} or removed from its hidden states to induce refusal behaviors \cite{arditi2024refusal}. However, the application of these powerful steering techniques in multilingual settings remains largely unexplored.
\section{Background}
\begin{table*}
    \setlength\tabcolsep{4pt} 
    \begin{center}
        
    \begin{tabular}{l|cccccccccc|c}
        \toprule
        \textbf{Methods} & \textbf{Es} & \textbf{Ja} & \textbf{Ru} & \textbf{Sw} & \textbf{Zh} & \textbf{Bn} & \textbf{Th} & \textbf{De} & \textbf{Fr} & \textbf{Te}  & \textbf{Avg}\\
        \midrule
      \rowcolor{lightgray!30} Llama2-7B\xspace & & & & & & & & & & & \\
            \quad Basic Prompt & 20.0 & 12.8 & 20.0 & .36 & 19.6 & 0.4 & 0.48 & 24.0 & 21.6 & 0.4  & 13.4 \\
        \quad Google-Trans & 26.4 & 24.4  & 24.8  & 26.0 & 27.6 & 26.0 & 24.0 & 22.4 & 24.4 & 24.0 & \textbf{25.0} \\
        \quad Self-Trans & 27.0 & 17.8 & 25.6 & 0.53 & 22.6 & 0.51 & 0.46 & 24.4 & 23.3 & 0.25 & 15.8 $\uparrow$\\
        \midrule
      \rowcolor{lightgray!30} Aya23-8B\xspace & & & & & & & & & & & \\
            \quad Basic Prompt & 40.0 & 25.6 & 34.4 & 0.64 & 27.6 & 1.0 & 13.1 & 36.0 & 32.0 & 0.16 & 22.7 \\
        \quad Google-Trans & 40.4 & 22.0 & 40.8 & 39.6 & 39.2 & 35.6 & 33.6 & 38.0 & 43.2 & 34.4 & \textbf{36.9}\\
        \quad Self-Trans & 33.6 & 25.6 & 27.8 & 0.52 & 22.0 & 10.6 & 16.6 & 34.6 & 33.2 & .01 & 21.0\\
        \midrule
    \end{tabular}
    \end{center}
    \vskip -0.3cm

    \caption{Comparison of Google-translated, native, and self-translated prompts on math tasks using LLaMA2-7B and Aya23-8B. \textbf{$\uparrow$} indicates improvement over the native prompt. Self-translation boosts LLaMA2 -7B by 2.4\%  and offers modest gains for Aya23, though both lag behind English performance.
}
    \label{tab:self-translate}
\end{table*}

\subsection{Evaluating LLM's capabilities}
Previous studies ~\cite{wendler2024llamas,zhao2024largelanguagemodelshandle}  indicate that LLMs often translate non-English prompts into English internally, which may limit their performance. To investigate this, a self-translation~\cite{etxaniz2023multilingual} process was used to assess whether LLMs understand non-English prompts or struggle with mistranslation. \autoref{tab:self-translate} shows that models like Llama2 ~\cite{touvron2023llama2openfoundation} and Aya23 \cite{aryabumi2024aya23openweight} can translate non-English tokens into English and that using this self-translation leads to a 2.4\% average improvement in Llama2's performance compared to native prompts. Aya23 also shows slight improvements for low-resource languages. However, the models still do not achieve the same level of understanding with non-English prompts as they do with English, likely due to representation mapping limitations.

\paragraph{Problem Formulation.}
Let a transformer model process a sequence of input tokens $t = (t_1, t_2, \ldots, t_n) \in \mathcal{V}^n$, producing a sequence of output probability distributions $y = (y_1, y_2, \ldots, y_n) \in \mathbb{R}^{n \times |\mathcal{V}|}$. Denote by $x_i^{(l)}(t) \in \mathbb{R}^d$ the residual stream activation at token position $i$ at the beginning of layer $l$. The residual stream is initialized via token embeddings: $x_i^{(1)} = \text{Embed}(t_i)$.\\
\noindent Each transformer layer updates the residual stream with attention and MLP components:
\begin{align}
    \tilde{x}_i^{(l)} &= x_i^{(l)} + \text{Attn}^{(l)}(x_1^{(l)}, \ldots, x_n^{(l)}) \\
    x_i^{(l+1)} &= \tilde{x}_i^{(l)} + \text{MLP}^{(l)}(\tilde{x}_i^{(l)}). \label{equation2}
\end{align}
After $L$ layers, the model computes the final logits $\text{logits}_i = \text{Unembed}(x_i^{(L+1)}) \in \mathbb{R}^{|\mathcal{V}|}$, followed by softmax to obtain the output distribution:
\begin{equation}
    y_i = \text{softmax}(\text{logits}_i) \in \mathbb{R}^{|\mathcal{V}|}.
\end{equation}

\paragraph{Activation Extraction and Alignment.}
We denote the post-layer-$L$ residual activations for English and a target language as $x^{(L+1)}_{\text{en}}$ and $x^{(L+1)}_{\text{target}}$, respectively. Our goal is to align the target language representations with their English counterparts by applying an additive transformation:
\begin{equation}
    x^{(L+1)}_{\text{altered}} = x^{(L+1)}_{\text{target}} + v^{(L)},
\end{equation}
where $v^{(L)}$ is a learned steering vector specific to layer $L$. We propose two methods for learning the alignment vector $v^{(L)}$:

\begin{enumerate}
    \item \textbf{Direct Preference Optimization (DPO).}
    Inspired by recent work, we apply Direct Preference Optimization (DPO) to learn $v^{(L)}$ by aligning target language representations with English ones while explicitly disaligning from original target activations. Unlike conventional approaches that compute the mean difference between activations~\cite{panickssery2024steeringllama2contrastive,wang2024trojanactivationattackredteaming} or use PCA to extract principal directions~\cite{dombrowski2023evaluating}, DPO learns a direction that better captures the bidirectional preference relationship between $x^{(L+1)}_{\text{en}}$ and $x^{(L+1)}_{\text{target}}$. This leads to improved multilingual alignment. See \autoref{steering_vector} for mathematical details.

    \item \textbf{Loss-Based Activation Alignment}
    Following the methodology ~\cite{park2023linear}, which suggests that representations in different languages may be linearly mappable, we also learn $v^{(L)}$ by minimizing the mean squared error between the aligned and English representations:
    \begin{equation}
        \mathcal{L}_{\text{MSE}} = \text{MSE}\left(x^{(L+1)}_{\text{en}}, x^{(L+1)}_{\text{altered}}\right).
    \end{equation}
    \end{enumerate}
\paragraph{Intervention.}
After learning the steering vector $v^{(L)}$, we apply an activation intervention by modifying the residual stream of a new target-language prompt at layer $L$. Specifically, given a new activation $x^{(L+1)}_{\text{target}}$ at layer $L+1$, we compute the intervened activation as:
\begin{equation}
    x^{(L+1)}_{\text{altered}} = x^{(L+1)}_{\text{target}} + v^{(L)}.
\end{equation}

\noindent This altered activation is then propagated through the remaining layers of the model, allowing us to observe how the intervention affects the model's output distribution. The goal is to steer the model’s internal representations of the target-language input to better align with English-like behavior.

\section{Experiments}
In this section, we outline the experimental setup necessary for the evaluations presented in Section \ref{sec:Exp_setup} and the corresponding results discussed in Section 4.2.

\subsection{Experimental Setup }
\label{sec:Exp_setup}
\textbf{Models}:~ We evaluated five prominent open-source models with varying levels of multilingual support: \textbf{LLama2-7B Chat}~\cite{touvron2023llama2openfoundation}, \textbf{Aya23-8B}~\cite{aryabumi2024aya23openweight}, \textbf{Gemma}~\cite{gemmateam2024gemmaopenmodelsbased}, \textbf{Qwen1.5 Chat}~\cite{qwen1.5}, and \textbf{LLama3-8B}~\cite{grattafiori2024llama3herdmodels}. For simplicity, the main discussion focuses on LLama2-7B Chat and Aya23-8B, while results for the remaining models are detailed in the appendix.\\\\
\textbf{Training Datasets}: To learn the steering vector, we used two datasets. For multilingual mathematical reasoning, we employed \textbf{MSVAMP}~\cite{chen2023breaking}, which spans 14 languages\footnote{es: Spanish, fr: French, ru: Russian, de: German, ja: Japanese, zh: Chinese, tr: Turkish, ar: Arabic, vi: Vietnamese, hi: Hindi, el: Greek, id: Indonesian, it: Italian, pt: Portuguese.} across high-, medium-, and low-resource tiers. For general tasks, we used the \textbf{Tatoeba} dataset~\cite{tiedemann-2020-tatoeba}, containing English–target language pairs across 50+ languages. We sampled 1,000 instances per language and grouped them by resource level to assess the effectiveness of our approach.
\\\\
\textbf{Evaluation Datasets}\label{evaluation_datast}: We evaluated our approach across five tasks spanning language understanding, commonsense reasoning, and mathematical reasoning: \textbf{MGSM}~\cite{shi2022language} for math, \textbf{XNLI}~\cite{conneau2018xnli} for natural language inference, \textbf{XCOPA}~\cite{ponti-etal-2020-xcopa} for causal commonsense, \textbf{MMLU}~\cite{hendrycks2020measuring} for general knowledge\footnote{We sampled 1k and 500 records from MMLU and XNLI, respectively.}, and \textbf{M3Exam}~\cite{zhang2023m3exammultilingualmultimodalmultilevel}, a human exam benchmark testing comprehensive language understanding. This diverse suite ensures a robust evaluation across linguistic competencies.
\\\\
\noindent \textbf{Baselines}: We compared seven baseline approaches for multilingual task handling:
\begin{itemize}
    \item \textbf{Basic Prompt}: The vanilla approach uses a traditional query format without any specialized prompting strategies.
    \item \textbf{Translate to English}: This method leverages LLMs’ strong English abilities by translating non-English inputs. Following \cite{liu2024translation}, we used two translation sources:
    \subitem \textbf{Google Translate}: A commercial service that translates examples into English.
    \subitem \textbf{NLLB} \cite{costa2022no}: An open-source model supporting over 200 languages.
    
    \item \textbf{XLT}~\cite{huang2023not}: A state-of-the-art prompting strategy that first translates the input question into English, then solves it step by step, leveraging the model’s stronger reasoning abilities in English.
    
    \item \textbf{5-shot Learning}~\cite{brown2020language}: Provides five examples to improve few-shot learning and multilingual generalization.
    \item \textbf{Supervised Fine-Tuning (SFT)}: This approach fine-tunes all model parameters on a non-English dataset and evaluates performance on downstream tasks.
\end{itemize}

\begin{table*}
\centering
\begin{tabularx}{\textwidth}{X|XXXXX}
\midrule
 \toprule
\textbf{\textit{Methods}} & \textbf{\textit{MGSM}} & \textbf{\textit{XCOPA}} & \textbf{\textit{XNLI}} & \textbf{\textit{M3EXAM}} & \textbf{\textit{MMLU}} \\ \hline

Base & 38.1 & 66.4 & 54.3 & 44.9 & 41.8 \\ 
Google trans & 44.6 & 75.5 & 57.1 & 49.8 & 48.7 \\ 
NLLB & 38.4 & 61.2 & 55.4 & 25.2 & 32.8 \\ 
5 shot & 32.5 & 58.9 & 41.3 & 29.2 & 30.2 \\ 
XLT & 30.4 & 46.1 & 51.7 & 37.7 & 30.1 \\ 
SFT & 34.7 & 55.2 & 46.1 & 39.9 & 35.4 \\ \midrule
DPO-Steer & $42.2_{+(4.1)}$ & $71.2_{+(4.8)}$ & $59.5_{+(5.2)}$ & $50.1_{+(5.2)}$ & $44.1_{+(2.3)}$ \\ 
MSE-Steer & $39.9_{+(1.8)}$ & $66.9_{+(0.5)}$ & $ 56.0_{+(1.7)}$ & $46.6_{+(1.7)}$ & $44.3_{+(2.5)}$ \\ \hline \midrule

\end{tabularx}

\caption{Average results across five open-source models in 14 languages. '+' indicates an improvement over the Base Prompts.}
\label{main_table}
\end{table*}

\vskip -2mm

\subsection{Results}

Our evaluation demonstrates in \autoref{main_table} that \textbf{activation-based steering is a highly efficient and scalable approach for improving multilingual language models}. Unlike resource-intensive methods such as Supervised Fine-Tuning (SFT), which require task-specific data, prolonged training, and careful hyperparameter tuning, our proposed techniques achieve competitive performance at a fraction of the computational and operational cost. Notably, DPO yields a substantial 26.7\% improvement over SFT, underscoring the effectiveness of targeted, activation-level interventions.

\noindent The advantages of this lightweight method are evident across a wide range of open-source baselines. Both DPO and MSE steering produce marked improvements over standard prompting strategies and even advanced cross-lingual transfer (XLT) prompts. The most significant gains are observed against a 5-shot in-context learning (ICL) baseline, where DPO achieves a 38.8\% improvement. \textbf{This result highlights steering's ability to correct internal representational misalignments that ICL, despite leveraging contextual examples, fails to resolve.} Further, DPO outperforms the open-source translation model NLLB by 25.4\%, demonstrating that steering is not merely a fine-tuning shortcut but a viable alternative to full translation pipelines. \textbf{It effectively aligns multilingual representations internally, without reliance on external systems.}  While steering does not yet exceed the performance of proprietary systems such as Google Translate, the margin is surprisingly narrow. DPO trails Google Translate by just 3.08\%, illustrating that internal, model-native interventions can approach the performance of large-scale production-grade translation APIs. This finding is particularly promising given the simplicity, interpretability, and deployability of the proposed steering method.

Across all experiments, \textbf{DPO consistently outperforms MSE-based steering.} We attribute this superiority to the directional optimization signal embedded in the DPO framework, which not only penalizes misalignment but actively guides the model toward improved representations. In contrast, the MSE objective quantifies error magnitude without providing gradient directionality, making optimization less efficient and less targeted. This fundamental distinction explains DPO’s effectiveness as a principled and robust method for steering multilingual behavior in pretrained language models. Moreover, the steering approach yields performance gains as model size increases, suggesting that larger language models benefit more from targeted activation interventions. We provide a detailed analysis of this trend in the \autoref{larger_size_sec}

\section{Analysis}
In this section, we analyze the proposed methods from various perspectives: Can we measure the quality of translation? What are the Challenges of steering? How transferable is the direction? And finally, which languages dominate the model's representation space?

\subsection{Can we quantify the quality of the internal translation process?}
\label{rq2}

\begin{figure}[t]
\small
  \includegraphics[width=0.99\columnwidth]{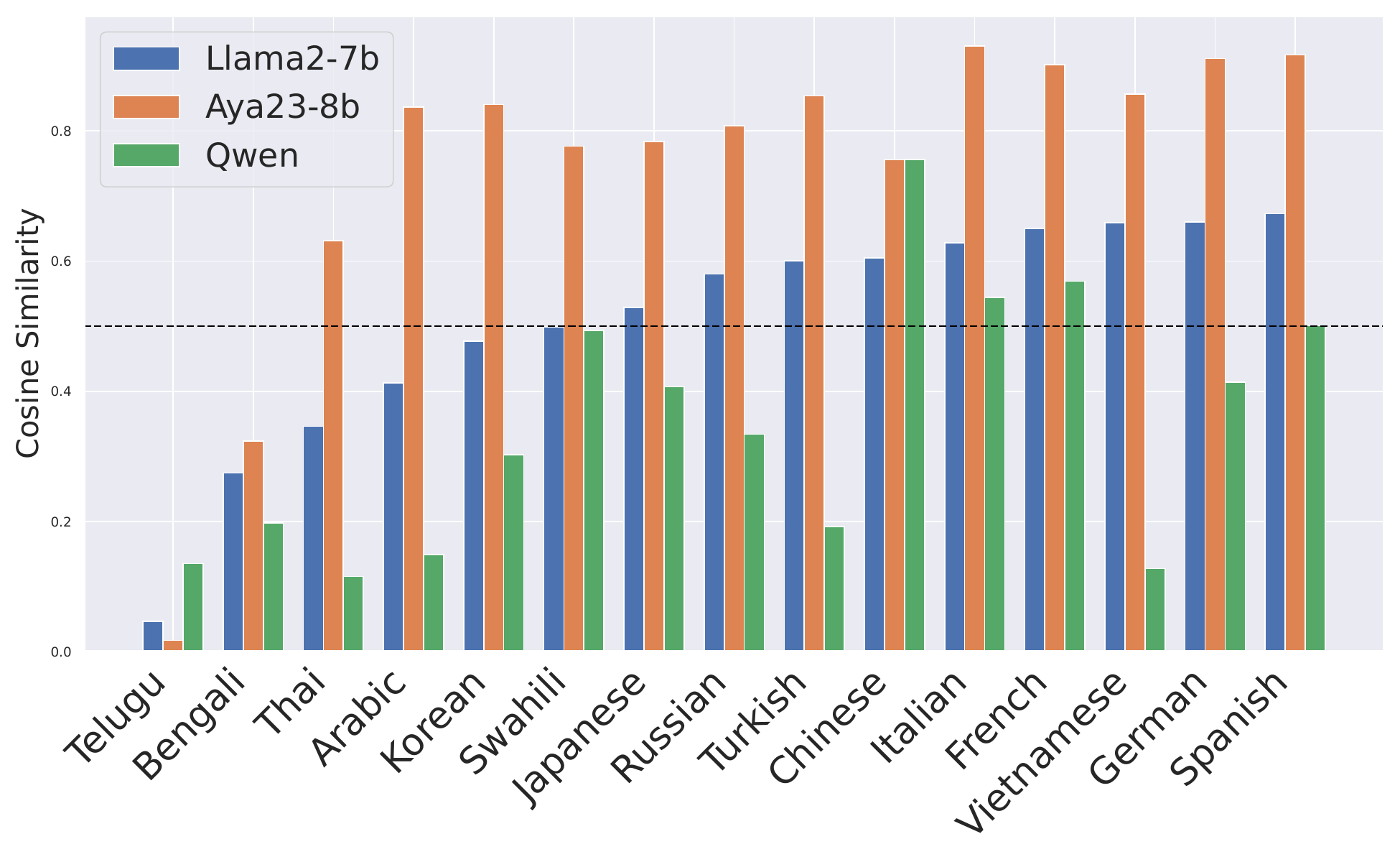}
      \vskip -0.3cm

  \caption{Similarity scores between language and English representations across models. The dashed line (threshold = 0.5) shows high-resource languages above and mid-/low-resource languages below it.}
  \label{fig:quanitfy_trans}
  \vskip -2mm
\end{figure}

Our analysis reveals a critical factor in a model's multilingual performance: the quality of its internal translation. When a model fails to accurately represent a language internally, it leads to information loss and significant performance gaps.
We quantify this internal translation quality by measuring how closely a language’s internal representation aligns with English, a proxy for how well it has been integrated into the model’s core space. Unsurprisingly, this alignment is directly tied to the volume of pre-training data for each language.
Models like\textbf{ LLaMA2 clearly illustrate this principle: high-resource languages with ample data (French, German)} show strong alignment with English, while low-resource languages (Thai, Telugu) exhibit much weaker connections. While specialized multilingual models like \textbf{Aya23 improve this alignment for some under-represented languages}, significant challenges remain see( \autoref{fig:quanitfy_trans})
In contrast, Qwen1.5 highlights a different risk: its training was so dominated by Chinese data that it often defaults to translating other languages into Chinese internally, struggling with alignment for most other languages.
Ultimately, these findings confirm that languages with weaker internal representations are fundamentally disadvantaged. This "representational misalignment" is not just a technical artifact; it is a direct cause of performance disparities across different language groups.

\subsection{Challenges in Steering Vector Generalization}

While effective, the steering vector approach faces two core limitations: it is distribution-sensitive and lacks contextual precision.

\noindent It is distribution-sensitive because its effectiveness depends heavily on the training distribution. As shown in \autoref{fig:fine_grained}, performance drops significantly on out-of-distribution tasks.
It lacks contextual precision because it applies a single, static correction across all inputs, ignoring prompt-specific variations. This uniform adjustment fails to capture the rich contextual variability inherent in natural language. These limitations suggest a clear direction for future work: developing context-aware steering mechanisms that adapt dynamically to each prompt, as explored by \cite{tran2025dynamic}.

\subsection{Is the steering vector transferable across languages?}
\label{rq3}
\begin{figure}[t]
    \small
  \includegraphics[width=0.99\columnwidth]{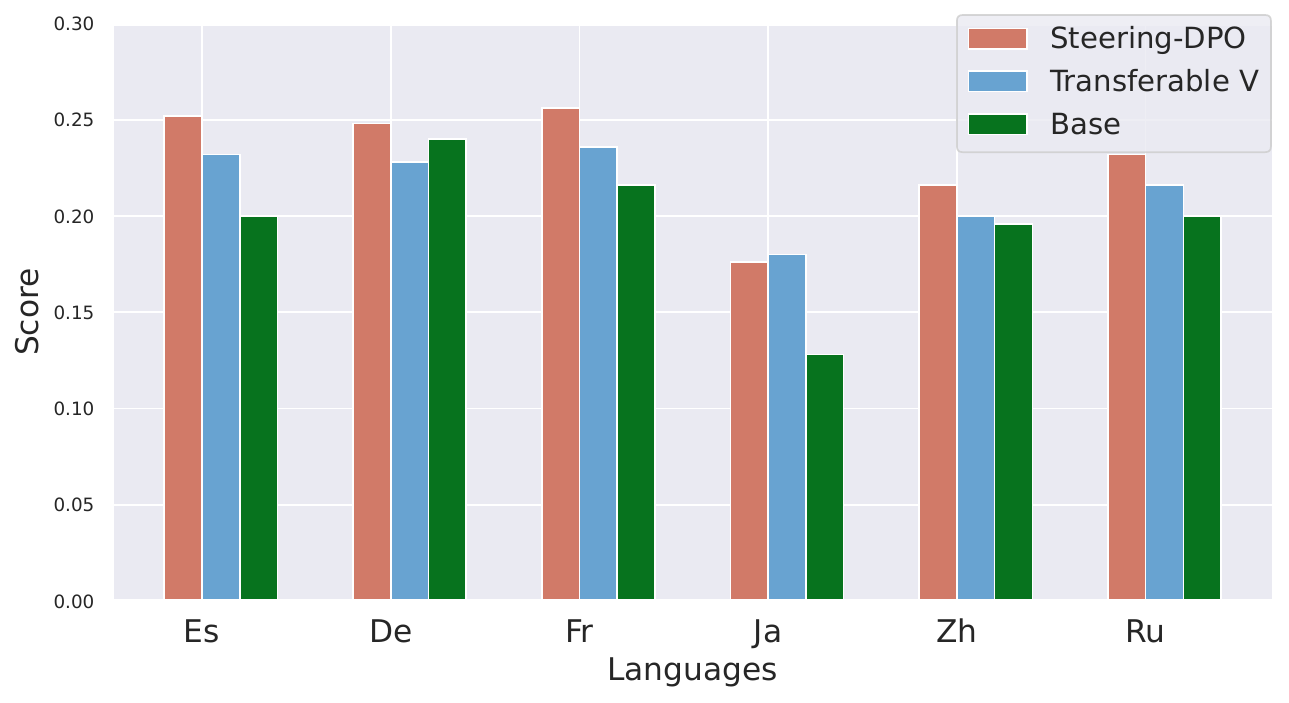}
      \vskip -0.3cm

\caption{Scores after applying steering vectors transferred directionally between language pairs (source → target), selected based on embedding similarity: Es→De, De→Es, Fr→Es, Ja→Zh, Zh→Ja, and Ru→Es.}
  \label{fig:transferability}
\vskip -4mm
\end{figure}

Building on prior work by \cite{cao2024personalized}, we examine whether a steering vector trained on one language can transfer effectively to another. Our results \textbf{indicate that transferability is feasible, but largely limited to languages within the same linguistic family, likely due to shared representational structures.}  As shown in \autoref{fig:transferability}, a vector trained on a source language consistently improves performance when applied to a related target language. For example, a vector trained on Spanish (Es) transfers well to German (De), French (Fr), and Russian (Ru) all Indo-European languages. Similarly, transfers between Japanese (Ja) and Chinese (Zh) are effective. However, these successes also expose the method's limitations. Cross-family transfers, such as from Spanish to Japanese, are ineffective, suggesting that while the vector captures more than language-specific patterns, it lacks the abstraction needed for generalization across distant language families.

\subsection{High Resource Languages are dominant in Representation Space}
A common assumption is that large language models "think" in English, merely translating other languages into English-like representations for processing. Our findings challenge this notion, revealing a more nuanced reality: the primary bottleneck is not English per se, but membership in a broader set of high-resource languages.
To investigate this, we selected three high-resource languages—Spanish, German, and French—chosen for their linguistic proximity and high representational similarity to English in the model. We then evaluated model performance on the MGSM reasoning task across these languages.
\autoref{tab:high_dominant_langs} shows that : performance in Spanish, German, and French \textbf{closely matched that of English, with negligible differences.}
These findings suggest that the \textbf{“English-centric}” view is overly reductive. Instead, current models appear to operate within a privileged set of high-resource languages capable of supporting complex reasoning. The central challenge for multilingual modeling is thus not merely accommodating many languages, but extending this inner circle to include low-resource languages.

\begin{table*}[h!]
\centering
\small
\renewcommand{\arraystretch}{1.5} 
\setlength{\tabcolsep}{8pt} 
\begin{tabularx}{\textwidth}{l l XXXXXX}
\midrule\midrule
\textbf{Models} & \textbf{Lang-Rep} & \textbf{Fr} & \textbf{Ru} & \textbf{Ja} & \textbf{Es} & \textbf{Zh} & \textbf{De} \\ \hline
\multirow{4}{*}{Aya23-8B} 
  & Fr & - & 34.3  & 25.6  & 40.0  & 27.6  &  36.0 \\ 
  & Es & 32.0 & 34.4  & 25.6  & -  & 27.6  & 36.0  \\ 
  & De & 32.0 & 34.4  & 25.6  & 40.0  & 27.6  & -  \\ \cmidrule(lr){2-8}
  & En & \textbf{38.0} & \textbf{41.2}  &  \textbf{34.8}  &  \textbf{44.4} & \textbf{32.8}  & \textbf{40.4}\\ \midrule \midrule
\multirow{4}{*}{Llama2-7B} 
  & Fr & - & 23.2  & 18.4  & 24.4  & 20.4  & 25.2  \\ 
  & Es & 24.4 & 22.8  & 17.6  &  - & 21.2  & \textbf{26.0}  \\ 
  & De & \textbf{26.0} & 21.6  & 17.6  & 24.4  & \textbf{22.0}  & -  \\ \cmidrule(lr){2-8}
  & En & 25.6 & \textbf{23.2}  & \textbf{20.8}  &\textbf{ 25.2}  & 21.6  & 24.8 \\
   \midrule\midrule
\end{tabularx}
\caption{The table highlights the selection of high-resource languages, such as French, Spanish, and German, as agnostic languages within the representation space of LLMs. The results indicate that English remains the most dominant language in this space. Other high-resource languages achieve comparable results, suggesting that their representations are distributed with similar likelihoods within the shared representation space.}
\label{tab:high_dominant_langs}
\end{table*}

\section{Ablation Studies}

\subsection{Impact of Injection Across Model Layers}

Our ablation studies reveal a critical insight: the optimal layer for steering is not universal. Instead, it is a direct reflection of a model's architecture and training data, as shown by the starkly different behaviors of Aya23 and LLaMA2.
\textbf{Aya23, a model explicitly designed for multilingualism, benefits most from steering in its early layers}. Its architecture is built to quickly translate different languages into a shared, language-neutral space. By correcting errors at this early stage—before they can cascade through the network—we see significant performance gains across all tasks, including mathematical reasoning (\autoref{fig:layers_tasks}).
In contrast, LLaMA2 presents a more complex picture. \textbf{For low-resource languages, steering the early and middle layers is highly effective}, as it helps these languages "catch up" and align with the model's dominant representations. However, for high-resource languages like English or German, this same intervention can be disruptive, interfering with already well-formed representations.
Finally, across both models, steering the final layers yields little to no improvement. This suggests that by this late stage, the model has already "committed" to its interpretation in its internal representation space. Intervening here is simply too late to have a meaningful effect. This confirms that to be effective, steering must happen "upstream," before the model's final reasoning process is complete.

\subsection{Impact of Steering Vectors on English Capabilities}
\noindent To assess the potential impact of steering vectors on the performance of monolingual English prompts, we evaluated nine different steering vectors, each tailored to a specific language and applied at various layers of the model. This evaluation aims to determine whether these vectors degrade the performance of English tasks, comparing the performance of each language-specific steering vector against the baseline monolingual results. Results in \autoref{tab:en_impact} \textbf{demonstrate that probing has a negative impact, which intensifies as the representational distance between two languages increases}. Conversely, the negative impact lessens for more similar languages. In models like LLaMA2, this correlation is pronounced, whereas, in Aya-23, which features more robustly represented languages, the impact is slightly reduced.

\begin{table}
  \centering
    \setlength\tabcolsep{2pt} 
    \small
    \begin{tabular}{lll}
        \toprule
        {Language} & {Llama2-7B} & Aya23-8B\\ \midrule
        {Es} & 31.6 & 42.0 \\ 
        {De} & 26.8 & 39.2 \\ 
        {Fr} & 26.4 & 41.2\\ 
        {Ja} & 24.8 & 40.8\\ 
        {Zh} & 25.6 &  41.6 \\ 
        {Ru} & 28.0 & 34.4 \\ 
        {Sw} & 26.8 & - \\ 
        {Bn} & 30.8 & -\\ 
        {Th} & 28.8 & - \\ \midrule
        {En} & \textbf{32.0} & \textbf{43.2} \\
        \midrule 
    \end{tabular}
    \vskip -0.3cm

    \caption{Results of MGSM task on Llama2-7B, Aya23-8B, the Steering vector has a negative impact on English Prompts.}
    \label{tab:en_impact}
\vskip -2mm
\end{table}

\section{Fine-tuning vs. Steering Approach}
\label{sec:sec_sft_steer}

Our findings suggest that \textbf{activation steering can achieve the same internal alignment benefits as fine-tuning,} but does so through a single, targeted intervention rather than a lengthy training process. To demonstrate this, we designed an experiment to visualize \textit{how} each method forces the model to align its internal representations with English.

\noindent We employed a "logit lens" analysis \cite{nostalgebraisty}, a technique that allows us to \textbf{peek inside the model}. At a specific layer \(L+1\), we take the model's internal state the post-activation output \(x^{(L+1)}_i\)~\ref{equation2} and project it back into the vocabulary space using the unembedding matrix. In simple terms, we ask the model: \textit{\textbf{"Based on your current state, what English word does this most look like?"}}
\[
\text{logits}^{(L+1)}_i = \text{Unembed}(x^{(L+1)}_i) \in \mathbb{R}^{|V|}
\]
By applying a softmax function to these logits, we can generate a probability distribution over the entire vocabulary, showing us the model's "best guess" for the next token.
\[
y_i = \text{softmax}(\text{logits}^{(L+1)}_i) \in \mathbb{R}^{|V|}
\]
\noindent We then use a language detection tool\footnote{\href{}{https://github.com/Mimino666/langdetect}}  on the most likely tokens to see if the model's internal state has successfully aligned with English. We compared three scenarios, with the full results shown in Figure~\ref{fig:sft_steer} (Appendix):
\begin{enumerate}
    \item \textbf{Base Model:} Exhibits weak cross-lingual alignment; target language representations remain distant from English.
    \item \textbf{Fine-Tuned Model:} Learns to align target and English representations after extensive translation fine-tuning.
    \item \textbf{Steered Model:} Matches this alignment instantly using a single steering vector.
\end{enumerate}

\noindent
\textbf{Both fine-tuning and steering improve representational alignment} with English: fine-tuning achieves this gradually over many steps, while steering provides an immediate, targeted correction.
\noindent
This efficiency raises a key question: \emph{do multilingual models still need steering?} Logit lens analysis shows that \textbf{advanced multilingual models already exhibit strong alignment with English, without requiring intervention.}
\noindent
Thus, the need for steering reflects limitations in base model training. While steering is an effective fix, better multilingual pretraining may eliminate the issue entirely. Full details are in Appendix~\ref{hyperparameters}.

\vskip -2mm
\section{Conclusions}

In this paper, we advance the study of multilingual processing in LLMs, exploring improvements across languages with varying resource levels. We analyzed LLM alignments from a multilingual perspective, highlighting how techniques like SFT and RLHF enhance multilingual capabilities by comparing these methods with steering and probing approaches and identifying limitations in steering vectors for handling linguistic nuances. Empirical experiments showed that probing inner layers boosts multilingual task performance but may hinder monolingual performance. Analysis of LLM families shows their sensitivity to layer-level changes, highlighting the importance of careful tuning and alignment to optimize multilingual performance.

\section*{Limitations}
We acknowledge that our approach, which involves probing by sweeping across all model layers, is not scalable for LLMs and is impractical for real-world applications. Moreover, the learnable steering vector is constrained by its fixed linear direction, limiting its capacity to capture the intricate mapping relationships between languages fully; learning steering vectors by individual tokens seems more promising than fixed steering. We leave this for future work. Additionally, Our experiments were also intentionally focused, isolating a single layer and language at a time. This controlled approach was necessary to establish a clear baseline, but a truly powerful system must learn to juggle multiple languages and layers simultaneously through more advanced, multi-objective training.

\section*{Ethics Statement}
This research adheres to ethical guidelines in the development and application of large language models (LLMs). We acknowledge the potential risks associated with multilingual processing, including biases in language representation, unequal performance across high- and low-resource languages, and the unintended consequences of steering techniques. Efforts were made to ensure transparency in our methodology and to mitigate biases by evaluating models across diverse languages and tasks. However, we recognize that our work may still reflect inherent biases present in the training data or model architectures. We encourage further research to address these limitations and promote equitable performance across all languages. Additionally, we emphasize the importance of responsible AI practices, including the careful deployment of LLMs in real-world applications to avoid harm or misuse.

\bibliography{custom}

\appendix
\onecolumn
\label{sec:appendix}

\section{Learning the Steering Vector}
\label{steering_vector}
In the first scenario, we utilize previous work \cite{cao2024personalized} that applied Direct Preference Optimization (DPO) methods to construct the steering vector. Specifically, Optimizing $v$ increases the probability of generating responses that align with the desired language behavior (e.g., English) while reducing the likelihood of responses associated with the opposite behavior (e.g., the target language). In this case, the contrast is defined between two language pairs: the English response $R_t$ and the target language response $R_O$.
\begin{equation} \label{eq:DPO}
     \min_v -\mathbb{E}_{d \sim \mathcal{U}, (q, r_T, r_O) \sim \mathcal{D}} \left[\log \sigma \left(d\beta \log \frac{\pi_{L+1}(r_T|A_L(q)+dv)}{\pi_{L+1}(r_T|A_L(q))} - d\beta \log \frac{\pi_{L+1}(r_O|A_L(q)+dv)}{\pi_{L+1}(r_O|A_L(q))}\right) \right].
\end{equation}
\\
Where: $v$ is the learnable steering vector, $\sigma$ represents the logistic function.$\beta$ controls the deviation from the original model. $\pi_{L+1}(\cdot|A_L(q))$ denotes the model's response from layer $L+1$, given the activation $A_L(q)$ at layer $L$ for the input question $q$.The term $d$ flips the optimization direction:
    \begin{itemize}
         \item  $d = 1$, the steering vector is optimized towards the English behavior $r_T$.
        \item If $d = -1$, the steering vector is optimized towards the opposite behavior $r_O$.  
    \end{itemize}
By optimizing this bi-directional objective, the steering vector $v$ is trained to align with either the desired target behavior or its reverse, depending on the directional coefficient d. This approach ensures that both language behaviors target and opposite are captured effectively, enhancing the model's ability to differentiate between them with precision.
\subsection{Algorithms}
\begin{algorithm}
  \caption{BiPO Steering Vector Learning}
  \label{alg:bipo}
  \begin{algorithmic}[1]
    \Require Pretrained LLM \textbf{$M$}, bilingual corpus $\mathcal{D}=\{(q_i, q_i^{\mathrm{en}})\}$, layer $L$, 
             learning rate $\eta$, epochs $T$
    \Ensure Steering vector $v \in \mathbb{R}^d$
    \State \textbf{Initialize} $v \leftarrow \mathbf{0}$
    \For{$e \leftarrow 1,\dots,T$}
      \ForAll{$(q, q^{\mathrm{en}})\in \mathcal{D}$}
        \State \Comment{1. Extract hidden activations at layer $L$}
        \State $h \leftarrow \mathrm{HiddenState}(M, q, L)$
        \State $h^{\mathrm{en}} \leftarrow \mathrm{HiddenState}(M, q^{\mathrm{en}}, L)$
        \State \Comment{2. Inject steering vector}
        \State $\tilde{h} \leftarrow h + v$
        \State \Comment{3. Compute logits from both activations}
        \State $\ell \leftarrow \mathrm{Logits}(M, \tilde{h})$
        \State $\ell^{\mathrm{en}} \leftarrow \mathrm{Logits}(M, h^{\mathrm{en}})$
        \State \Comment{4. Direct Preference Optimization (DPO) loss}
        \[
          \mathcal{L}_{\mathrm{DPO}}
          \;=\;
          -\mathbb{E}_{t\sim\mathcal{V}}\bigl[\log\sigma\bigl(\ell^{\mathrm{en}}_t - \ell_t\bigr)\bigr]
          \quad\text{(see App.\,A, eq.\ref{eq:DPO})}
        \]
        \State \Comment{5. Gradient‐step update}
        \State $v \leftarrow v \;-\; \eta\,\nabla_v\,\mathcal{L}_{\mathrm{DPO}}$
      \EndFor
    \EndFor
    \State \textbf{Return} $v$
  \end{algorithmic}
  \label{DPO_algorithm}
\end{algorithm}

\begin{algorithm}[H]
  \caption{MSE Steering Vector Learning}
  \label{alg:mse}
  \begin{algorithmic}[1]
    \Require Pretrained LLM $M$, bilingual corpus $\mathcal{D}=\{(q_i, q_i^{\mathrm{en}})\}$, layer $L$, 
             learning rate $\eta$, epochs $T$
    \Ensure Steering vector $v \in \mathbb{R}^d$
    \State \textbf{Initialize} $v \leftarrow \mathbf{0}$
    \For{$e \leftarrow 1,\dots,T$}
      \ForAll{$(q, q^{\mathrm{en}})\in \mathcal{D}$}
        \State \Comment{1. Extract hidden activations at layer $L$}
        \State $h \leftarrow \mathrm{HiddenState}(M, q, L)$
        \State $h^{\mathrm{en}} \leftarrow \mathrm{HiddenState}(M, q^{\mathrm{en}}, L)$
        \State \Comment{2. Inject steering vector}
        \State $\tilde{h} \leftarrow h + v$
        \State \Comment{3. Compute Mean‑Squared Error loss}
        \[
          \mathcal{L}_{\mathrm{MSE}}
          \;=\;
          \frac{1}{d}\,\big\|\tilde h \;-\; h^{\mathrm{en}}\big\|_2^2
        \]
        \State \Comment{4. Gradient‑step update}
        \State $v \leftarrow v \;-\; \eta\,\nabla_v\,\mathcal{L}_{\mathrm{MSE}}$
      \EndFor
    \EndFor
    \State \textbf{Return} $v$
  \end{algorithmic}
\end{algorithm}
\subsection{Other learning methods}
\begin{figure*}[h]
\begin{center}
\includegraphics[width=.99\textwidth]{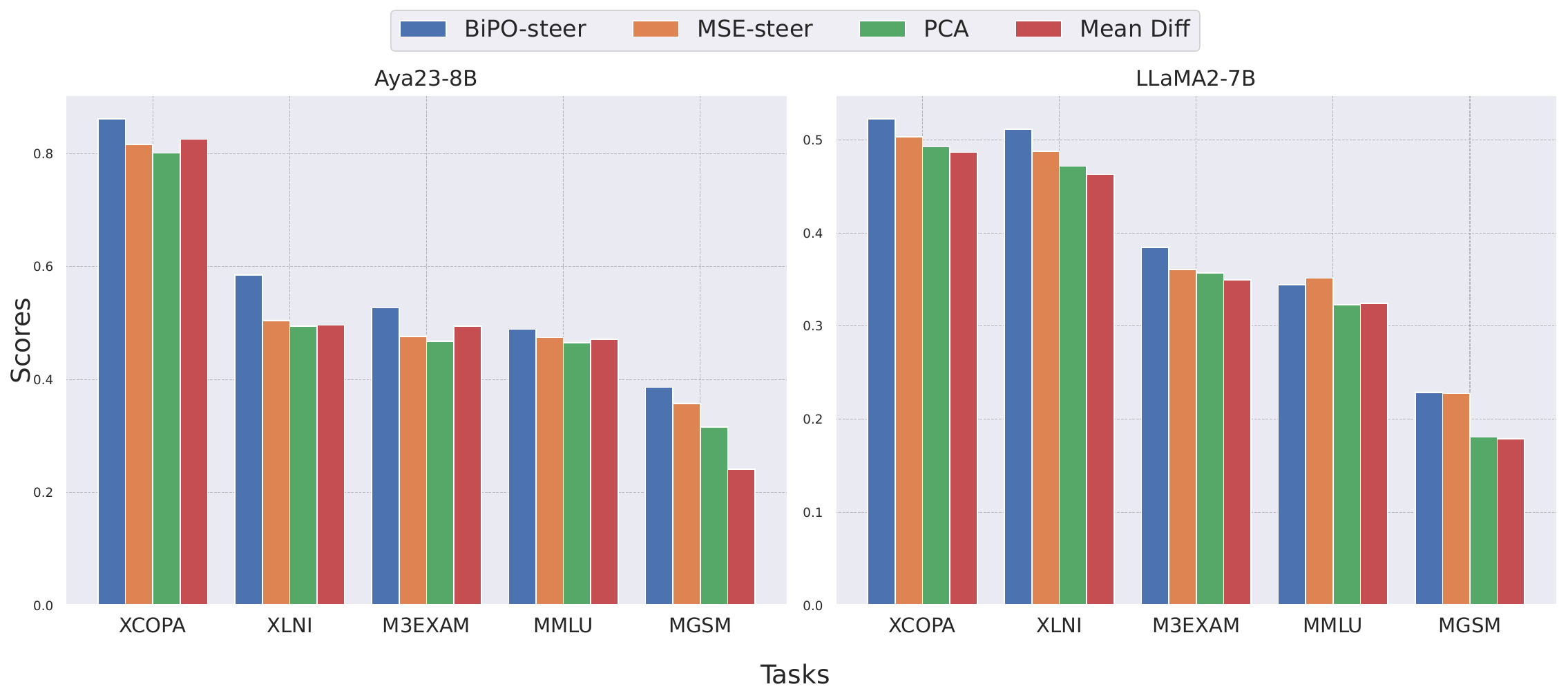} 
\caption{
demonstrate that using learnable steering vectors surpasses PCA and the Mean Difference approaches across all tasks on two models: Aya32-8B and LLama2-7B.}
\label{fig:learn_vector}
\end{center}

\end{figure*}

Effectively learning a manifold that encapsulates the feature representations between languages is vital for bridging the distributional gap across linguistic boundaries. While prior approaches~\cite{cao2024personalized,zou2023representation}, such as PCA and calculating the mean difference between constructive activations (CAA), have been shown to shift activation distributions, they fall short in accurately capturing essential features in multilingual contexts. In contrast, advanced methods like BiPO excel by leveraging a dynamic feedback loop during the manifold learning process, enabling them to better align multilingual representations.~\autoref{fig:learn_vector} highlights the performance of various models across diverse tasks, underscoring the effectiveness of this approach.

\section{SFT Vs Steering: Problem Setup and Notation}
\label{SFT_steer}

Let $\mathcal{M}$ represent the base LLM and $\mathcal{M}^*$ denote the fine-tuned version trained on an instruction dataset $\mathcal{D}$, where $\mathcal{D} = {(Q_i, A_i)}_{i=1}^n$ consists of question-answer pairs. To analyze the mechanisms of fine-tuning, we model the transformation of each layer $l$ as:
\begin{equation}
H_{l}(x) = h_{l}(x) + S_{l}(x)
\end{equation}
where:
\begin{itemize}
\item $h_{l}(x)$ represents the original layer $l$ activation for input $x$
\item $S_{l}(x) \in \mathbb{R}^{d}$ is a learnable parameter matrix that modulates the activation in the residual stream
\item $d$ is the dimensionality of the hidden state
\end{itemize}
For each $(Q, A) \in \mathcal{D}$, $H_{l}$ is optimized via the loss function:
\begin{equation}
\mathcal{L}(\mathcal{M}^(Q), A) = -\sum_{t=1}^T \log P(a_t|a_{<t}, Q; \theta^)
\end{equation}
where:
\begin{itemize}
\item $\theta^*$ represents the fine-tuned model parameters
\item $a_t$ is the $t$-th token in the answer $A$
\item $T$ is the length of the answer
\end{itemize}                   
In contrast, the steering approach learns a single steering vector $v \in \mathbb{R}^d$ that modifies activations across all layers:
\begin{equation}
H_{l}(x) = h_{l}(x) + \alpha v
\end{equation}
where $v$ is the learned steering direction , $\alpha$ is a scaling coefficient that controls the magnitude of steering ,The same $v$ is applied across different $(Q, A)$ pairs
\section {Hyperparameters}
\label{hyperparameters}
\textbf{Training Steering Vectors}: For all models, we followed the authors'~\cite{cao2024personalized} configurations, setting \( \beta = 0.1 \), using the AdamW optimizer with a learning rate of \( 5 \times 10^{-4} \), and applying a weight decay of 0.05. The batch size was set to 1, and we utilized a cosine learning rate scheduler with 100 warmup steps. The number of epochs was set to 1 for all models, except for certain languages in LLama2 and Aya23-8B, where it was increased to 3 epochs. For the MSE method, we used a learning rate of \( 1 \times 10^{-8} \) and varied the number of epochs in the range [3, 5, 8, 12]. Mean Squared Error (MSE) was used as the loss function, and cosine similarity was employed to evaluate the similarity between raw activations during training.\\ \\
For the supervised fine-tuning described in \autoref{sec:sec_sft_steer}, we trained the models on the same training datasets for 5 epochs, using a learning rate of  \( 1 \times 10^{-3} \), a weight decay of 0.001, and a warmup ratio of 0.05. The batch size was set to 16, and we utilized a cosine learning rate scheduler with the AdamW optimizer.

\subsection{High-Capability Models and Inner Translation Behavior}

\begin{figure*}
\begin{center}
\includegraphics[width=.99\textwidth]
{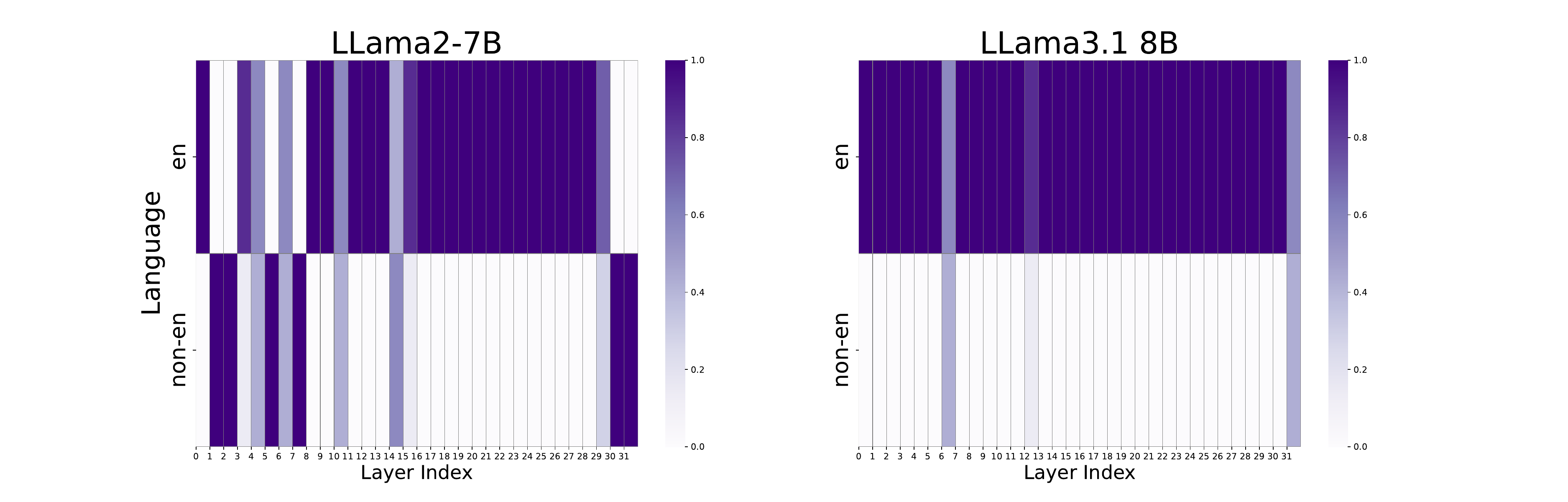} \\
\includegraphics[width=.99\textwidth]{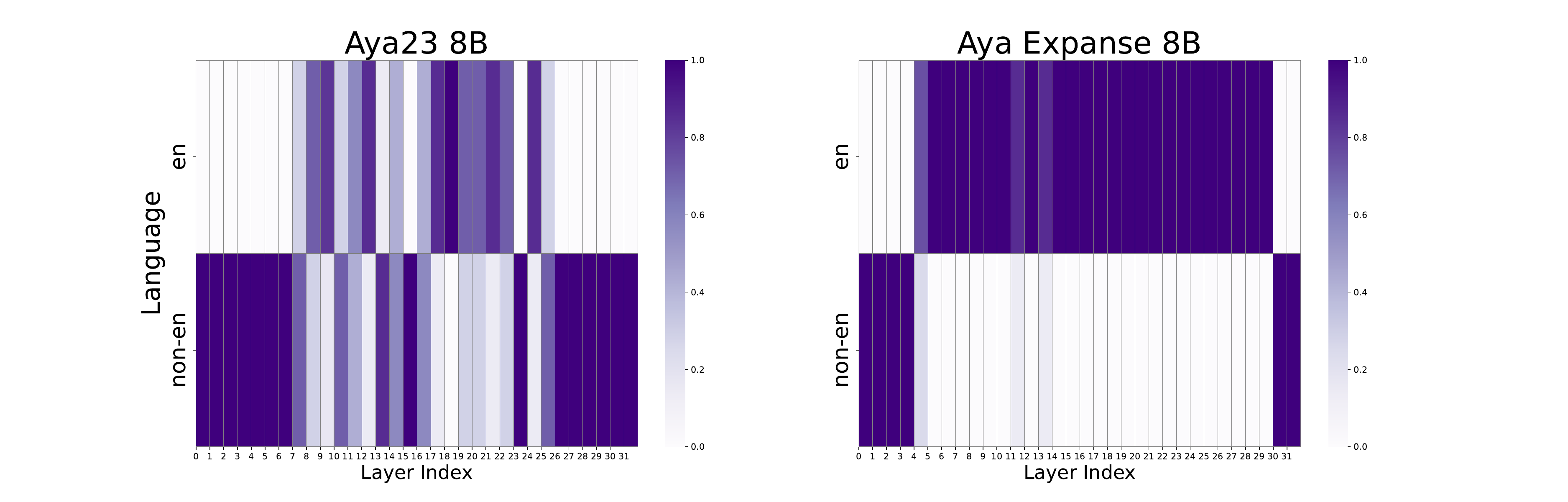}

\caption{
illustrates the processing of multilingual tokens in models of varying capabilities within the same family. LLama3.1 demonstrates a strong alignment of tokens into English-aligned representations, whereas LLama2 struggles with this. Similarly, Aya-Expanse exhibits robust token alignment, attributed to RLHF techniques, while Aya23 shows weaker alignment.}
\label{fig:high_capable_models}
\end{center}

\end{figure*}

In this section, we investigate the behavior of high-capacity multilingual LLMs, such as LLama3.1~\cite{grattafiori2024llama3herdmodels} and Aya23-Expanse~\cite{odumakinde2024multilingual}, to understand the factors behind their superior performance across languages. Using the logit lens, we analyze their internal representations and find that multilingual processing primarily occurs in the initial layers, with minimal inner translation loss (illustrated in \autoref{fig:high_capable_models}). These models map multilingual representations onto an English-aligned distribution early on, creating a shared, agnostic space. This alignment, enhanced by techniques like SFT and reinforcement RLHF, explains their effectiveness. For instance, Aya-Expanse shows significant improvements due to these methods~\cite{dang2024rlhf}. Our findings align with prior studies, confirming that SFT and RLHF substantially boost multilingual performance, consistent with earlier observations on the impact of SFT on internal representations~\cite{dang2024rlhf}.

    

\section{Larger LLMs Exhibit Consistent Behavior}
\label{larger_size_sec}
To address translation loss misalignment in larger language models, we extended our evaluation of steering approaches to larger architectures. Due to computational constraints, we tested only LLama2-13B on the MGSM task. \autoref{tab:llama2-13b} indicates that these larger models follow the same trend of performance improvements across different languages, mirroring the behavior observed in smaller models.
\begin{table}
    \setlength\tabcolsep{5pt} 
    \begin{center}
        
    \begin{tabular}{l|cccccc|c}
        \toprule
        \textbf{\textit{MGSM}} & \textbf{Es} & \textbf{Fr} & \textbf{Ru}  & \textbf{De} & \textbf{Ja} & \textbf{zh}  & \textbf{Avg}\\
        \midrule
      \rowcolor{lightgray!30} \textit{Llama2-13B}\xspace & & & & & & &   \\
            \quad Basic Prompt & 33.6 & 30.0 & 28.0 & 30.8 & 18.0 & 26.4 & 27.8 \\
        \quad Google-Tr & 39.2 & 35.2 & 36.8 & 36.4 & 35.6 & 36.4 & 36.6 \\
        \quad NLLB & 35.2  & 33.6 & 32.0 & 34.0  & 20.0 & 28.0 & 30.4  \\
        \quad 5$@$shots & 35.2 & 32.8 & 26.8 & 33.2 & 18.4 & 23.6 &  28.3 \\
        \quad XLT & 33.6  & 30.4 & 30.8 & 27.6 & 25.2 & 29.6 & 29.5  \\
        \quad SFT & 35.4 & 35.0 & 31.8 & 34.4 & 26.0 & 28.1 & 31.7   \\  \midrule
        \midrule
         Bipo-method & $36.8_{(+3.2)}$ & $33.2_{(+3.2)}$ & $31.6_{(+3.6)}$ & $35.2_{(+4.4)}$ & $26.8_{(+8.8)}$ & $29.2_{(+2.8)}$ & $32.1_{(+4.3)}$  \\
         MSE-method &  $32.4_{(-1.2)}$ & $34.8_{(+4.8)}$ & $34.0_{(+6)}$ & $35.2_{(+4.4)}$ & $24.4_{(+6.4)}$ & $30.0_{(+3.6)}$ & $31.8_{(+4.0)}$ \\ \midrule \midrule
     
    \end{tabular}
    \end{center}

    \caption{Results of the MGSM Task Evaluated on the Llama2-13B Model Across Diverse Languages}
    
    \label{tab:llama2-13b}
\end{table}

\label{more_figures}

\begin{figure*}[h]
\begin{center}
\includegraphics[width=0.99\textwidth]{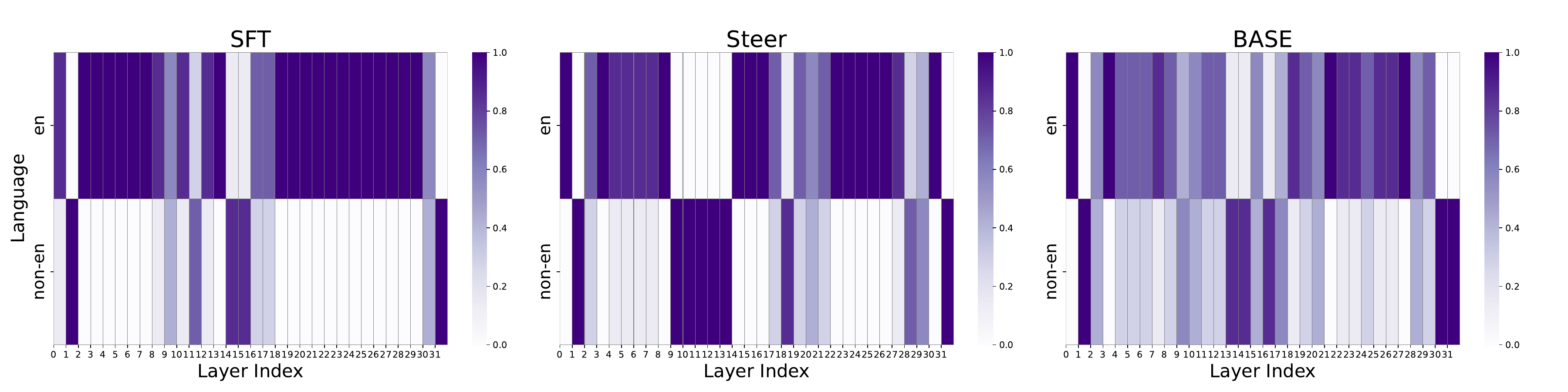}
  \caption{: Distribution of non-English token values across model layers at three different stages: pre-fine-tuning (base model), post-fine-tuning(SFT), and after applying steering at a specific layer. The results demonstrate that both fine-tuning and steering exhibit similar behavior, aligning non-token values more closely with English token distributions.
}
  \label{fig:sft_steer}
\end{center}
\end{figure*}

\begin{figure*}
\begin{center}
\includegraphics[width=.99\textwidth]
{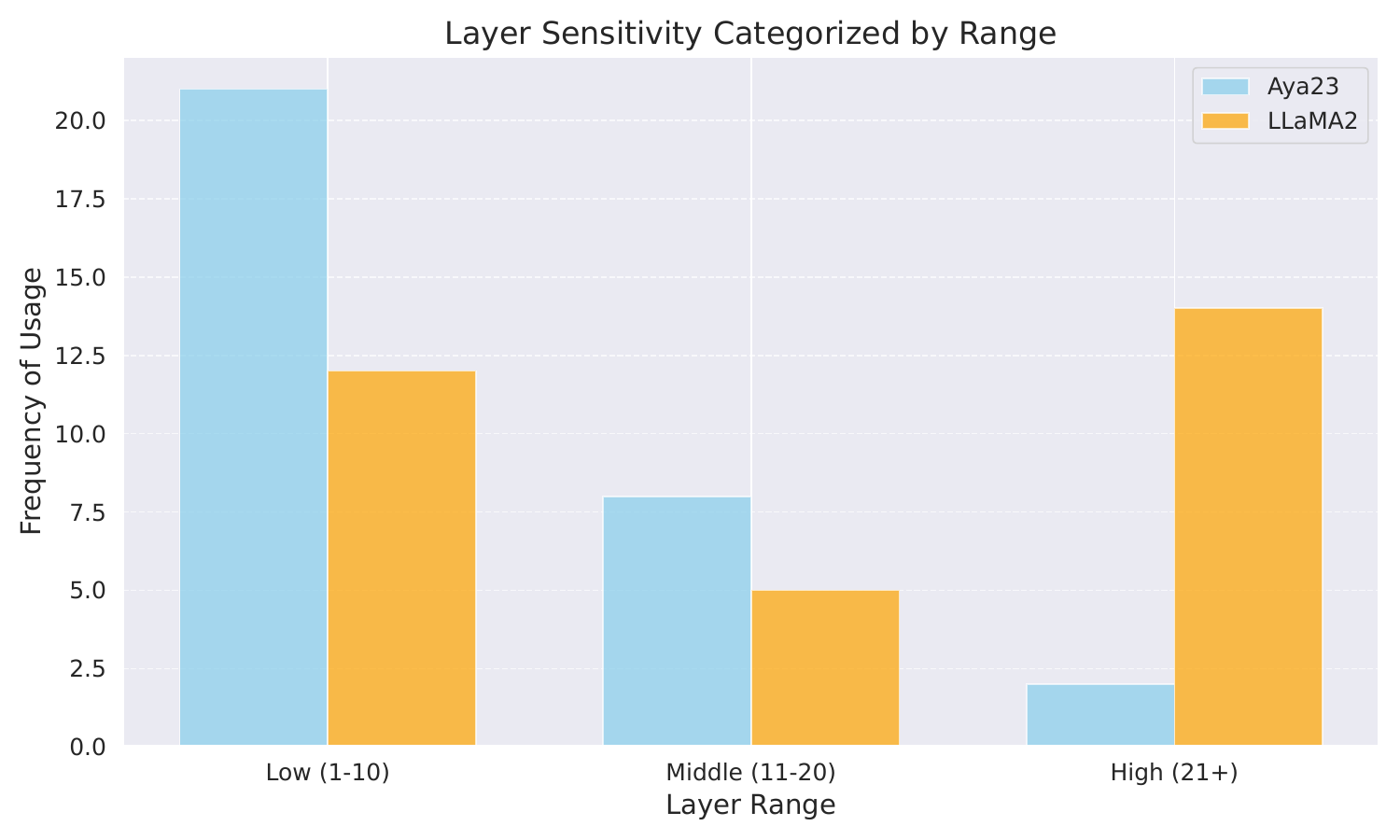} 

\caption{
illustrates the layers most sensitive to probing across two models. Aya23 demonstrates high sensitivity in the initial layers but exhibits reduced performance in the middle and later layers. In contrast, LLama2 experiences a notable drop in performance in the middle layers, with improved results in the later layers. Additionally, the initial layers of LLama2 perform better for low- and medium-resource languages.}
\label{fig:layers_tasks}
\end{center}

\end{figure*}

\begin{figure*}
\subfloat[XNLI]{\includegraphics[width = 3in]{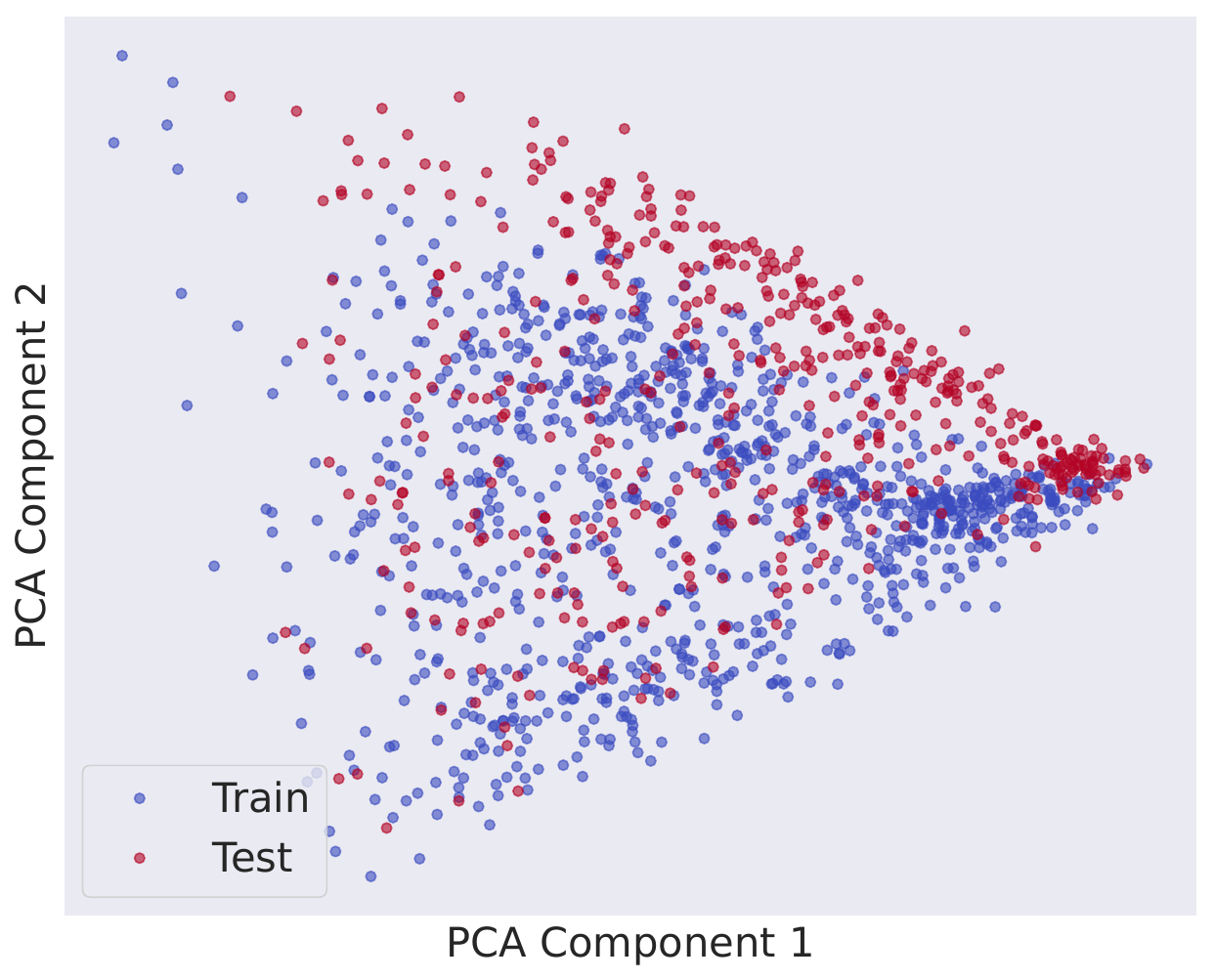}} 
\subfloat[MGSM]{\includegraphics[width = 3in]{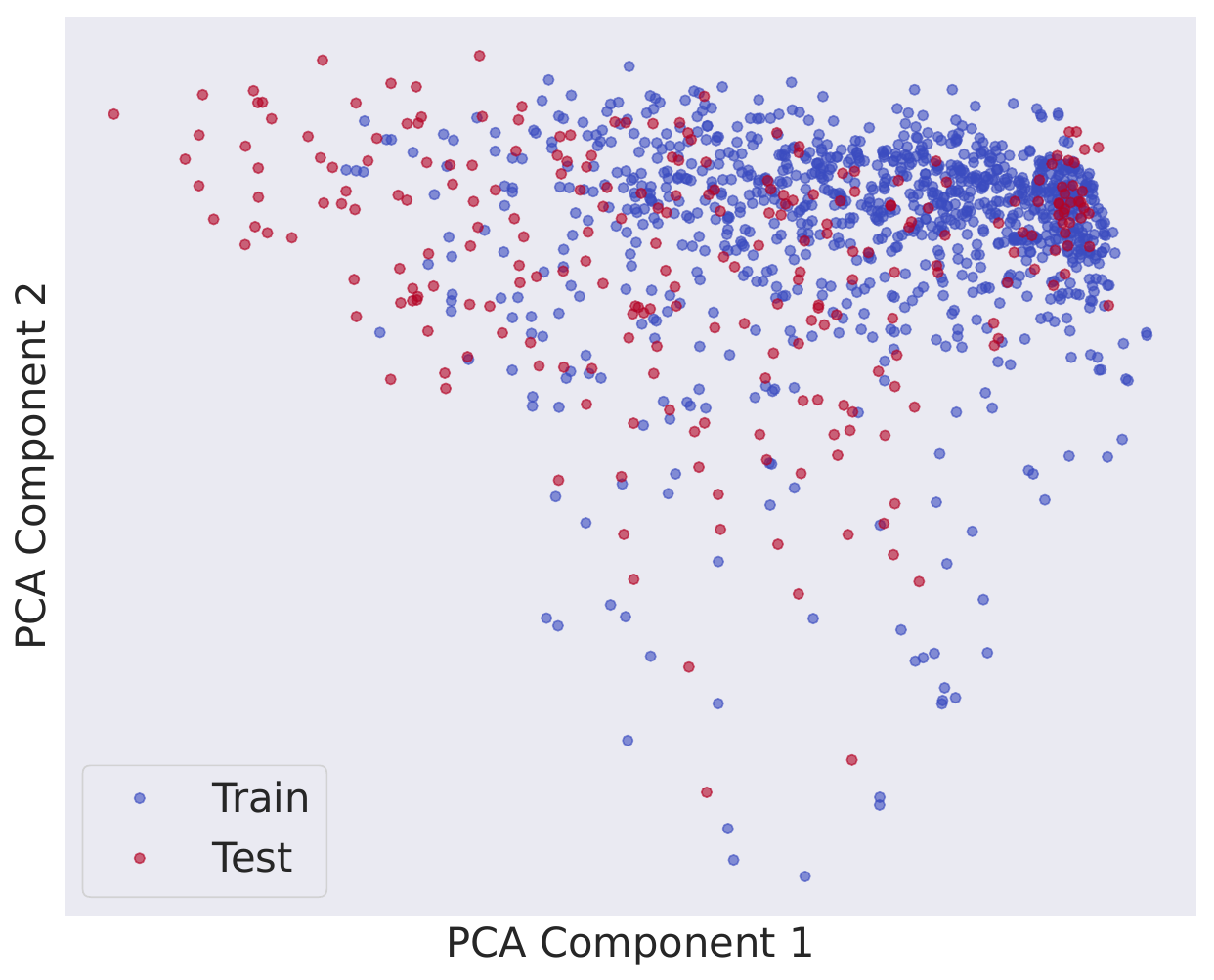}}\\
\subfloat[M3EXAM]{\includegraphics[width = 3in]{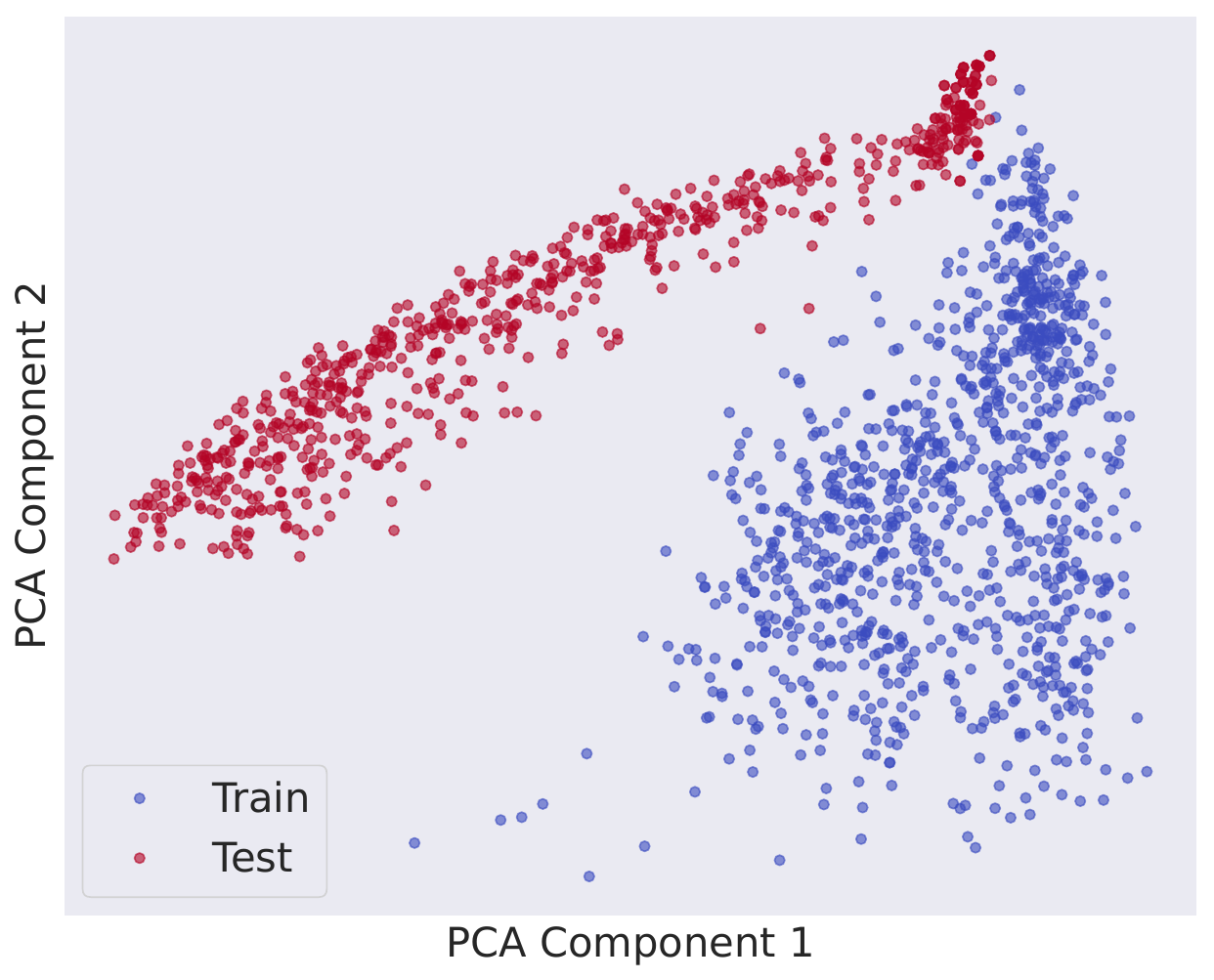}} 
\subfloat[MMLU]{\includegraphics[width = 3in]{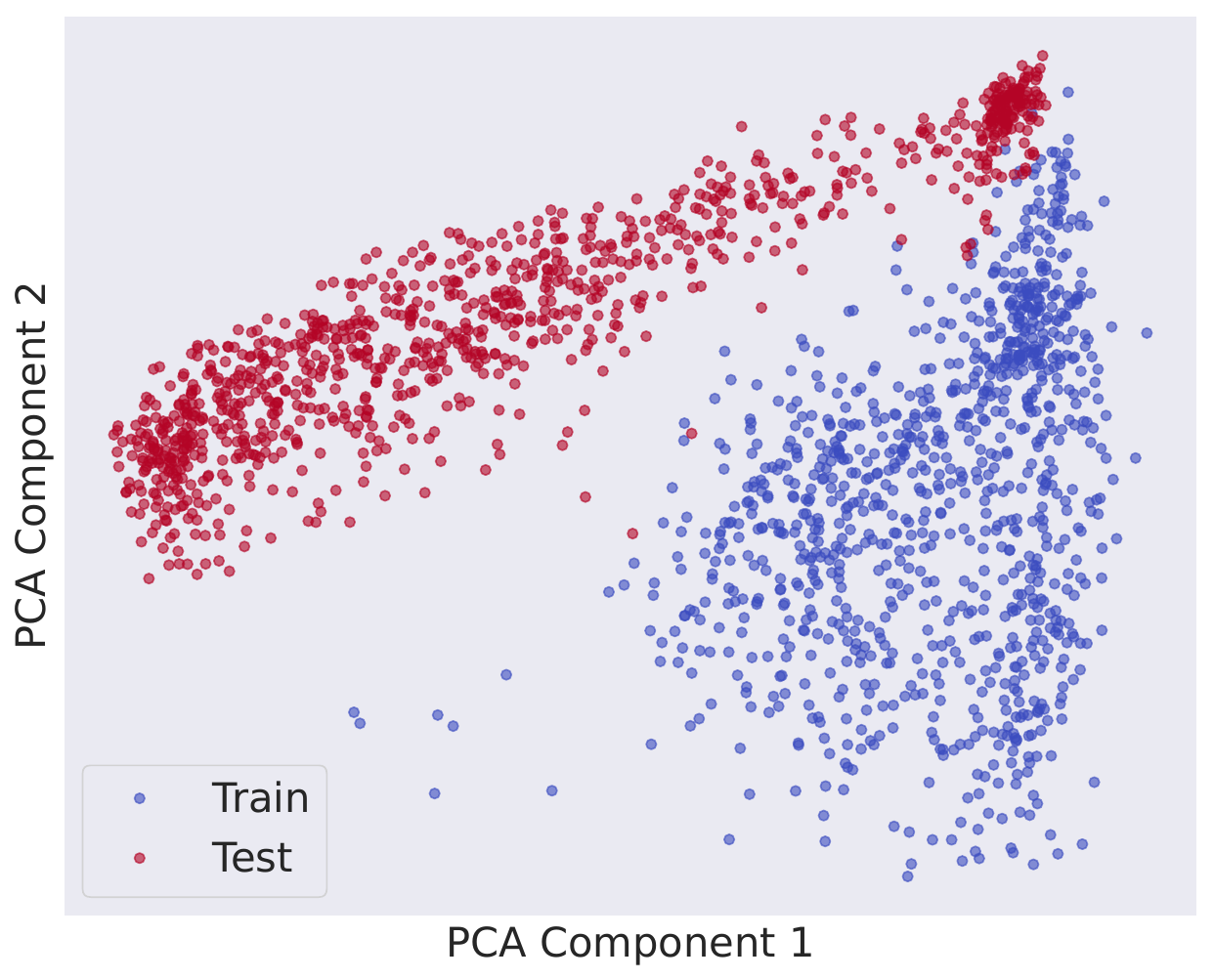}}
\\

\caption{
The sub-figures illustrate the distribution of the training and testing datasets across various tasks, emphasizing that steering approaches perform effectively when the testing dataset's distribution closely aligns with the training dataset's distribution but show limited improvement when the two distributions differ significantly.}
\label{fig:fine_grained}

\end{figure*}


\begin{table*}[h]
\centering
\small

\begin{tabularx}{\textwidth}{X|XXXXX}
\midrule\midrule
\multicolumn{6}{c}{\textbf{\textit{Aya32-8B}}}\\
\midrule
 \toprule
\textbf{\textit{Methods}} & \textbf{\textit{MGSM}} & \textbf{\textit{XCOPA}} & \textbf{\textit{XNLI}} & \textbf{\textit{M3EXAM}} & \textbf{\textit{MMLU}} \\ \hline

\noindent Basic & 32.6 & 81.6 & 49.9 & 46.9 & 45.3  \\ 
Google-Tr & 37.6 & 83.9 & 52.4 & 49.4 & 50.7 \\ 
NLLB & 32.3 & 73.2 & 49.9 & 26.5 & 34.0 \\ 
5@Shot & 36.1 & 84.5 & 59.8 & 42.5 & 30.9 \\ 
XLT & 26.9 & 12.1 & 52.7 & 38.8 & 27.0 \\ 
SFT & 34.4 & 82.0 & 49.8 & 47.4 & 46.0   \\  \midrule
DPO-Steer & 38.6 & 86.1 & 58.5 & 52.7 & 49.0 \\
MSE-steer & 35.7 & 81.6 & 50.5 & 47.4 & 47.6 \\  \midrule\midrule
\multicolumn{6}{c}{\textbf{\textit{Llama2-7B}}} \\ \midrule\midrule
Basic & 19.6 & 47.6 & 46.9 & 30.6 & 31.3 \\ 
Google-Tr & 25.0 & 51.8 & 50.9 & 42.8 & 41.5 \\ 
NLLB & 22.6 & 40.4 & 49.7 & 20.9 & 24.5 \\ 
5@Shot & 12.2 & 29.6 & 14.7 & 12.6 & 24.4  \\
XLT & 20.2 & 47.2 & 45.8 & 28.5 & 23.6   \\  
SFT & 24.0 & 49.9 & 49.4 & 36.4 & 34.0   \\  \midrule
DPO-Steer & 22.9 & 52.2 & 55.1 & 38.4 & 34.4 \\ 
MSE-steer & 22.8 & 50.3 & 48.7 & 35.1 & 36.0 \\  \hline \midrule
\multicolumn{6}{c}{\textbf{\textit{Llama3-8B}}} \\ \midrule\midrule
Basic & 62.0 & 66.7 & 63.2 & 51.6 & 50.7 \\ 
Google-Tr & 70.7 & 79.3 & 65.8 & 54.5 & 58.2 \\ 
NLLB & 60.0 & 63.4 & 63.4 & 23.9 &  40.7 \\ 
5@Shot & 55.6 & 63.5 & 27.6 & 24.1 &  26.0 \\ 
XLT & 26.9 & 56.9 & 55.0 & 39.2 & 33.7 \\ 
SFT & 64.7 &  72.2 & 63.9 & 53.8 & 51.6   \\  \midrule
DPO-Steer & $67.0_{(+5.0)}$& $75.0_{(+8.3)}$ & $64.3_{(+1.1)}$ & $ 55.3_{(+3.7)}$ & $52.8_{(+2.1)}$\\
MSE-steer & $62.8_{(+0.8)}$  & $68.4_{(+1.7)}$ & $64.0_{(+0.8)}$ &  $53.0_{(+1.4)}$ & $50.6_{(-0.1)}$  \\  \midrule\midrule
\multicolumn{6}{c}{\textbf{\textit{Gemma-7B}}} \\ \midrule\midrule
Basic & 27.3 & 66.2  & 46.4 & 37.3  & 39.6 \\ 
Google-Tr & 37.4 & 83.1  & 51.0 & 45.4 & 47.0 \\ 
NLLB & 29.8 & 65.4 & 50.0 & 23.0 & 33.8 \\ 
5@Shot & 12.2 & 42.2 & 39.6 & 20.2 & 22.0 \\
XLT & 28.7 & 49.8 & 49.9 & 28.1 & 26.5   \\
SFT & 28.6 &  67.8 & 49.2 & 43.1 & 40.8   \\  \midrule
DPO-Steer & $30.0_{(+2.7)}$ & $68.8_{(+2.6)}$  & $ 51.9_{(+5.5)}$ & $45.7_{(+8.4)}$ & $41.1_{(+1.5)}$ \\ 
MSE-steer & $27.6_{(+0.3)}$ & $68.0_{(+1.8)}$  & $47.9_{(+1.5)}$ & $37.5_{(+0.2)}$ & $41.1_{(+1.5)}$ \\ 
\midrule\midrule
\multicolumn{6}{c}{\textbf{\textit{Qwen1.5-7B}}} \\ \midrule\midrule
Basic & 49.2 & 70.1  & 66.0 & 58.3  & 42.1 \\ 
Google-Tr & 52.6 & 79.3  & 65.4 & 57.2 & 46.1 \\ 
NLLB & 47.4 & 63.4 & 64.3 & 31.8 & 31.3 \\ 
5@Shot & 46.8 & 74.6 & 65.0 & 46.7 & 48.1 \\
XLT & 49.4  & 64.4 & 55.6 & 54.2 & 39.7   \\ 
SFT & 50.6 &  71.6 & 67.4 & 58.2 & 43.8   \\  \midrule
DPO-Steer & $ 52.8_{(+3.6)}$ & $73.8_{(+3.7)}$  & $67.8_{(+1.8)}$ & $58.7_{(+0.4)}$ & $42.9_{(+0.8)}$ \\ 
MSE-steer & $50.5_{(+1.3)}$ & $70.7_{(+.7)}$  & $ 68.7_{(+2.7)}$ & $59.7_{(+1.4)}$ & $46.4_{(+4.3)}$ \\ 
\midrule\midrule
\end{tabularx}

\caption{Presents the average accuracy across various tasks for Three models: Llama3-8B , gemma-7Ba and Qwen1.5-7B . The evaluation spans over 10 languages, covering high, medium, and low-resource levels. (+) symbol indicates the average improvement compared to the Basic Prompt baseline.}
\label{tab:other_models}
\end{table*}

\clearpage
\section{Examples of different tasks}Examples of model-generated responses under both original and steered conditions when prompted with questions in different languages. The correct answers are highlighted in \textcolor{green}{\textbf{green}}, while incorrect answers are marked in \textcolor{red}{\textbf{red}} for clarity.

\begin{CJK}{UTF8}{mj}

\begin{table}[h]
\resizebox{1.0\linewidth}{!}{
    \begin{tabular}{ l | p{1.3\linewidth} }
    \toprule
      \textbf{Question} & マンディはベネディクトに\$100借りがあります。彼らは毎月2\%の利子がつくことに同意しました。マンディが3か月後に返済できるとすると、彼女はベネディクトにいくら払えばいいですか？ Let's think step by step. \\
      \midrule
       \textbf{Original generation}  & Mandy borrowed \$100 from Benedict with an agreement to pay 2\% interest per month. \\
       & 1. Calculate the total interest due after 3 months: \\
       & \$100 x 2\% x 3 months = \$6 \\
       & So, the total amount Mandy owes Benedict after 3 months is \$106 (\$100 + \$6). \\
       & 2. Calculate the monthly payment amount: \\
       & To find the monthly payment amount, we need to divide the total amount owed by the number of months left to repay the loan. In this case, there are 3 months left to repay the loan, so: \\
       & \$106 ÷ 3 months = \$35.33 \\
       & So, Mandy needs to pay \textcolor{red}{\textbf{\$35.33}} per month to Benedict. \\
      \midrule
      \textbf{Steered generation}  & 1. Mandi borrowed \$100 from Benedict with an agreement to pay 2\% interest per month. \\
      & 2. The loan is to be repaid 3 months from now. \\
      & So, we need to calculate the total amount that Mandi needs to pay to Benedict to settle the loan. \\
      & First, let's calculate the interest for the 3 months: \\
      & 2\% per month x 3 months = 6\% interest \\
      & Now, let's calculate the total amount that Mandi needs to pay: \\
      & Original amount borrowed: \$100 \\
      & Interest: \$6 (6\% of \$100) \\
      & Total amount to be paid: \$106 \\
      & So, Mandi needs to pay \textcolor{green}{\textbf{\$106}} to Benedict to settle the loan. \\ 
      \bottomrule \bottomrule
    \end{tabular}
    }
\end{table}
    \vspace{0.5cm}

\begin{table}[h]
\resizebox{1.0\linewidth}{!}{
   \begin{tabular}{ l | p{1.3\linewidth} }
    \toprule

      \textbf{Question} & Les canes de Janet pondent 16 œufs par jour. Chaque matin, elle en mange trois au petit déjeuner et en utilise quatre autres pour préparer des muffins pour ses amis. Ce qui reste, elle le vend quotidiennement au marché fermier, au prix de 2 \$ l'œuf de cane frais. Combien (en dollars) gagne-t-elle chaque jour au marché fermier ? Let's think step by step. \\
      \midrule
       \textbf{Original generation}  & Janet's chickens lay 16 eggs per day. \\
       & She eats 3 eggs for breakfast every day. \\
       & She uses 4 eggs to make muffins for her friends. \\
       & So, the number of eggs she has left to sell at the farmer's market is: \\
       & 16 eggs/day - 3 eggs/day = 13 eggs/day \\
       & Since she sells each egg at \$2, Janet earns: \\
       & 13 eggs/day x \$2/egg = \$26/day \\
       & So, Janet earns \textcolor{red}{\textbf{\$26}} per day at the farmer's market. \\
      \midrule
      \textbf{Steered generation}  & 1. Janet lays 16 eggs per day. \\
      & 2. She eats 3 eggs for breakfast every day. \\
      & 3. She uses 4 eggs to make muffins for her friends. \\
      & 4. So, how many eggs does she have left to sell at the farmer's market? \\
      & 16 eggs/day - 3 eggs/day = 13 eggs/day \\
      & 13 eggs/day - 4 eggs/day (for muffins) = 9 eggs/day \\
      & Janet has 9 eggs left to sell at the farmer's market every day. \\
      & Now, let's calculate how much money she makes from selling those eggs: \\
      & 9 eggs/day x \$2/egg = \$18/day \\
      & So, Janet makes \textcolor{green}{\textbf{\$18}} per day selling eggs at the farmer's market. \\ 
      \bottomrule \bottomrule
    \end{tabular}
    }
\end{table}
    \vspace{0.5cm}

\begin{table}[h]
\resizebox{1.0\linewidth}{!}{
   \begin{tabular}{ l | p{1.3\linewidth} }
    \toprule
        \textbf{Question} & A seguir estão questões de múltipla escolha sobre Linguagens. Por favor, dê apenas a opção correta, sem quaisquer outros detalhes ou explicações. \\
      & \textbf{Por que as formigas não morrem quando postas em forno de micro-ondas?} \\
      & As micro-ondas são ondas eletromagnéticas com frequência muito alta. Elas causam vibração nas moléculas de água, e é isso que aquece a comida. Se o prato estiver seco, sua temperatura não se altera. Da mesma maneira, se as formigas tiverem pouca água em seu corpo, podem sair incólumes. Já um ser humano não se sairia tão bem quanto esses insetos dentro de um forno de micro-ondas superdimensionado: a água que compõe 70\% do seu corpo aqueceria. Micro-ondas de baixa intensidade, porém, estão por toda a parte, oriundas da telefonia celular, mas não há comprovação de que causem problemas para a população humana. \\  
      & OKUNO, E. Disponível em: http://revistapesquisa.fapesp.br. Acesso em: 11 dez. 2013. \\  
      & Os textos constroem-se com recursos linguísticos que materializam diferentes propósitos comunicativos. Ao responder à pergunta que dá título ao texto, o autor tem como objetivo principal: \\
      & A. defender o ponto de vista de que as ondas eletromagnéticas são inofensivas. \\
      & B. divulgar resultados de recentes pesquisas científicas para a sociedade. \\
      & C. apresentar informações acerca das ondas eletromagnéticas e de seu uso. \\
      & D. alertar o leitor sobre os riscos de usar as micro-ondas em seu dia a dia. \\
      & E. apontar diferenças fisiológicas entre formigas e seres humanos. \\
      & Responder:\\
      \midrule
       \textbf{Original generation}  & A resposta correta é a alternativa: \textcolor{red}{\textbf{E}}. \\
      \midrule
      \textbf{Steered generation}  & The correct answer is \textcolor{green}{\textbf{C}}. \\ 
      \midrule
      \end{tabular}
      }
      \end{table}
    \vspace{0.5cm}

\end{CJK}

\end{document}